\documentclass[fleqn,10pt]{SelfArx} % Document font size and equations flushed left

\usepackage[english]{babel} % Specify a different language here - english by default
\usepackage{caption}
\usepackage{subcaption}
  \hypersetup{pdfauthor={Kamer Vishi},
     pdftitle={Attack Analysis of Face Recognition Authentication Systems Using Fast Gradient Sign Method},
     pdfsubject={Journal Article},
     pdfkeywords={Biometrics, Adversarial Attack, Authentication, Face Recognition, FGSM, Machine Learning, Targeted Attack, Untargeted, Security, Privacy, IAM, Machine Learning, Subjective Logic, Deep Learning, Deepfake, Spoofing, Cyber Attacks, Insider Threats, Quantum Cryptography},
     pdfpagelayout={TwoPageLeft}, % Displays two pages, odd-numbered pages to the left 
     pdfcreator={Xelatex},
     colorlinks={true},  % Colors the text of links and anchors. 
     anchorcolor={black},%Color for anchor text.
     filecolor={cyan}, %Color for URLs which open local files.
     menucolor={red},%Color for Acrobat menu items.
     runcolor={blue} %Color for run links (launch annotations).
 }           

\definecolor{color1}{RGB}{0,0,90} % Color of the article title and sections
\definecolor{color2}{RGB}{0,20,20} % Color of the boxes behind the abstract and headings

\usepackage{hyperref} % Required for hyperlinks

\hypersetup{
	hidelinks,
	colorlinks,
	breaklinks=true,
	urlcolor=color2,
	citecolor=color1,
	linkcolor=color1,
	bookmarksopen=false,
	pdftitle={Title},
	pdfauthor={Author},
}

\JournalInfo{This is an Accepted Manuscript of an article published by Taylor \& Francis in Applied Artificial Intelligence on September 16, 2021. Available online: \href{https://www.tandfonline.com/doi/full/10.1080/08839514.2021.1978149}{doi:10.1080/08839514.2021.1978149}} % Journal information
\Archive{} % Additional notes (e.g. copyright, DOI, review/research article)
\PaperTitle{Attack Analysis of Face Recognition Authentication Systems Using Fast Gradient Sign Method} % Article title

\Authors{Arbena Musa\textsuperscript{1}, Kamer Vishi\textsuperscript{2}, and Blerim Rexha\textsuperscript{1}*} % Authors
\affiliation{\textsuperscript{1}\textit{Faculty of Electrical and Computer Engineering, University of Prishtina, Prishtina, Kosovo}} % Author affiliation
\affiliation{\textsuperscript{2}\textit{Department of Informatics, University of Oslo, Oslo, Norway}} % Author affiliation
\affiliation{*\textbf{Corresponding author}: Blerim Rexha - Email: blerim.rexha@uni-pr.edu} % Corresponding author

\Keywords{Biometrics -- Adversarial Attack -- Authentication -- Face Recognition -- FGSM -- Machine Learning -- Targeted Attack -- Untargeted} % Keywords - if you don't want any simply remove all the text between the curly brackets
\newcommand{\keywordname}{Keywords} % Defines the keywords heading name

\Abstract{Biometric authentication methods, representing the ”something you are” scheme, are considered the most secure approach for gaining access to protected resources. Recent attacks using Machine Learning techniques demand a serious systematic reevaluation of biometric authentication. This paper analyzes and presents the Fast Gradient Sign Method (FGSM) attack using face recognition for biometric authentication. Machine Learning techniques have been used to train and test the model, which can classify and identify different people’s faces and which will be used as a target for carrying out the attack. Furthermore, the case study will analyze the implementation of the FGSM and the level of performance reduction that the model will have by applying this method in attacking. The test results were performed with the change of parameters both in terms of training and attacking the model, thus showing the efficiency of applying the FGSM.}

\begin{document}
\maketitle % Output the title and abstract box
%\tableofcontents % Output the contents section
\thispagestyle{empty} % Removes page numbering from the first page
%----------------------------------------------------------------------------------------
%	ARTICLE CONTENTS
%----------------------------------------------------------------------------------------
\section*{Introduction} % The \section*{} command stops section numbering

\addcontentsline{toc}{section}{Introduction} % Adds this section to the table of contents
% no \IEEEPARstart
The transition of most physical processes in the form of the electronic application, starting from personal data storage to banking, has forced the requirement of authentication whenever you need to access them. Almost every modern computer or mobile device offers a large number of authentication forms to increase security. Emphasizing that face recognition is one of the most common forms of authentication that we use efficiently daily, attacking processes that use this way of authentication is dangerous. The range in which facial recognition is used starts from tagging photos on social networks and opening personal devices to access high-security areas. The risk of malfunctioning facial recognition systems can be severe and significant damage. As attacks on authentication processes are becoming more and more inevitable, work is actively underway to identify vulnerabilities in the models created for this process and improve the vulnerabilities that have already been made known in advance.

According to \cite{Andress2011}, authentication, in terms of information security is the set of methods we use to verify whether an identity is genuine. It should be noted that authentication only determines whether the claim of the instance's identity is correct. This is not related to the access offered subsequently to the authenticated party and what it is allowed to do because it falls under the responsibility of the authorization process. Several methods can be used for authentication that is categorized by reference to a factor. In trying to validate an identity claim the more factors used, the more positive the results are. When we authenticate, factors can be something we know, something we are, or something we have. Authentication is the act of proving that someone is whom they claim to be. By focusing on something we are, we can categorize the relatively unique physical attributes of an individual, such as biometric characteristics.

Biometrics summarizes measurements and calculations of human characteristics. Biometric authentication is used in computer science as a form of identification and verification. Biometric identifiers are distinctive, measurable features used to label and describe individuals. Biometric identifiers are often categorized as physiological or behavioral characteristics.
\begin{figure*}[htb]
     \centering
\includegraphics[width=0.95\textwidth]{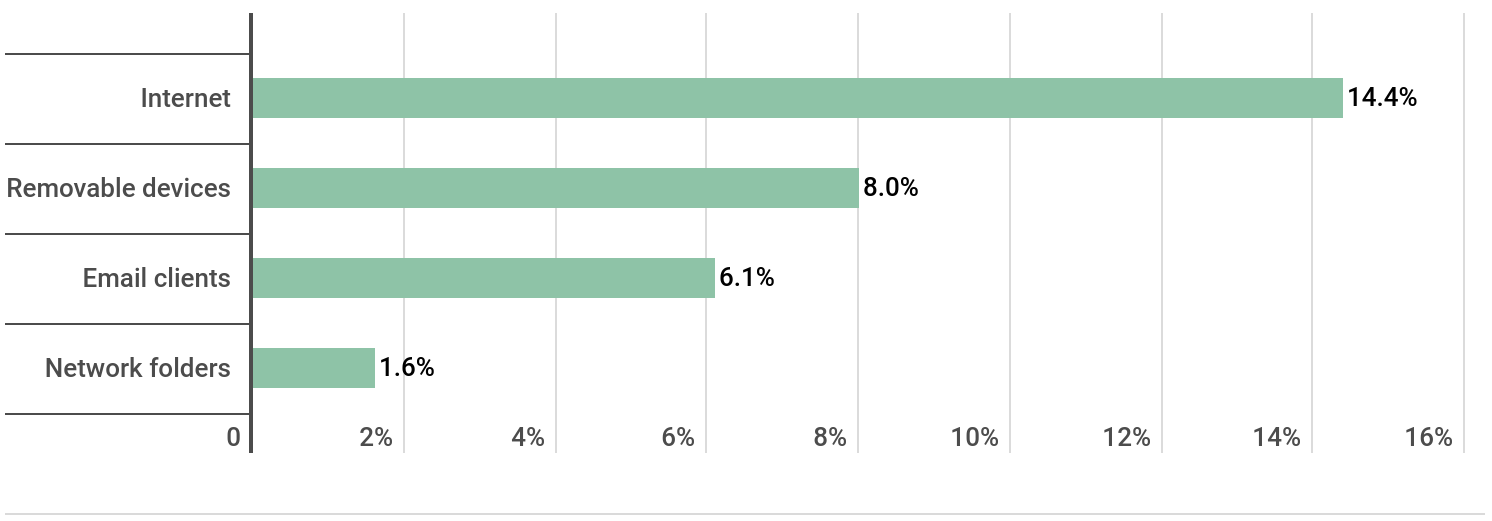}
     \caption{Threats for biometric data processing and storage systems \cite{Kaspersky2019}}
     \label{fig:biometrics_threats}
\end{figure*}
A biometric system recognizes a person by determining authentication by using different features of his body such as fingerprints, face, iris, retina scanning, signature, hand geometry, voice, etc. These features can be divided based on the level of security, user acceptance, cost, performance, etc. Biometrics is a growing technology, which has been widely used in forensics, access and physical security \cite{Rubal2019}.

Despite the benefits, biometrics is prone to some security threats, such as presentation attacks (spoofing), biometric data processing attacks, software, and networking vulnerabilities. According to \cite{Kaspersky2019}, in 2019, $14.4\%$ of all threats on biometric data processing systems were internet-related threats (malicious and phishing websites), followed by removable media ($8\%$) and network folders ($6.1\%$), as presented in Figure \ref{fig:biometrics_threats}.

Being aware of the importance of the authentication process and the consequences of its performance may have, any attack carried out on the process is a threat to many instances that exploit it. If the airport authentication system was attacked, people wanted by the police could be allowed to travel without any problems. In cases where authentication is used in automotive, a vehicle could be used by unauthorized persons or even cause improper functioning by confusing driver's preferences. In high-security buildings, any malfunction of authentication systems could lead to unauthorized persons having access to areas they are not supposed to be in. In schools, the safety of students could be endangered. In banking systems there is the possibility of card theft and misuse of large sums of money by deceiving an identity identification system.

We chose FGSM because it is a white-box approach, as it requires access to the internals of the classifier being attacked. As mentioned in \cite{Xue_2019}, FGSM image perturbation algorithm is to disturb images at the classification layer, and it needs to be considered that the size of the disturbance factor is determined, resulting in the false classification effect of face recognition is not obvious. In addition, FGSM is compared with other attack methods in \cite{Rao2020ATC}. They use the area under curve (AUC) as an evaluation metric. The value of AUC ranges from 0 to 1. A perfect model would approach 1 while a poor model would approach 0. For single models, FGSM has higher AUC than other methods such as PGD, MIFGSM, DAA, DII-FGSM, in the white box attacks.

The contribution of this paper is towards of using human images with FGSM, which represents an alternative approach to other traditional methods mentioned previously (PGD, MIFGSM, DAA, DII-FGSM).

Therefore the objectives of this study are:
\begin{itemize}
  \item Measuring the performance of a model based on different training characteristics.
  \item Implementing attack with the FGSM method.
  \item Determining the level of vulnerability of the model depending on the training characteristics.
  \item Determining the effectiveness of the attack on the model with different performances.
\end{itemize}
The rest of the paper is structured as follows. Section~\ref{sec:relatedwork} gives an overview of related work, Section~\ref{sec:fgsm} describes the Fast Gradient Sign Method (FGSM), Section~\ref{sec:experimental_setup} presents the experimental setup and data analysis. Experimental results are provided in Section~\ref{sec:results}, and lastly, Section~\ref{sec:conclusion} rounds off the paper with a discussion and indication of future work.

%------------------------------------------------
\section{Related Work}\label{sec:relatedwork}
%\textcolor{red}{In the following subsections we will give the related works..}

\subsection{Face Recognition}
Since the face is considered to be the most important part of the human body, it plays a key role in identifying and validating a person, due to which it can be used for multiple applications in daily life. As studied in \cite{Zulfiqar2019} the face recognition system is trending across the globe as it offers safe and reliable security solutions. It is the fastest biometric technology that identifies a person without involving him. It does not add any unwanted delay as it automatically captures the image of a person from a certain distance or takes a snippet from a video, processes that image, and recognizes that person. From \cite{Mori2005}, \cite{Yong2020} it can be seen that for the creation of a system that performs the function of face recognition, ML techniques are used, specifically artificial neural networks. Neural networks that have been created specifically for such tasks and that have shown efficiency in performing them when it comes to image processing, analysis, and comparison are convolutional neural networks (CNN) \cite{Albawi2017}. This type of network was also used in the case of our study.

\subsection{Transfer Learning}
While using Machine Learning, if a model is trained with a certain dataset and labels, it responds only to the classification of images in the respective labels. With the increasing use of models and the extremely large addition of domains treated as training subjects of these models, training from the beginning becomes costly and unnecessary to be repeated. From this came the creation of the concept of Transfer Learning. From the name itself, it is understood that the transferred learning is related to the transfer of prior knowledge acquired by a trained model with a certain dataset, to another model that will be trained with a different dataset but which belongs to the same domain. The main pillar of the trained model used in this study is precisely transfer learning. Deep Learning and Convolutional Machine Learning algorithms are designed to work in isolation. These algorithms are trained to solve specific tasks and models must be constructed from the beginning when the spatial distribution of features changes. Transfer learning is the idea of overcoming the paradigm of isolated learning and utilizing the knowledge gained while solving one task to solve another complicated task. Some of the architectural types of CNN networks developed in advance, which are used for the further development of AI models are: LeNet \cite{LeNet}, AlexNet
\cite{AlexNet}, VGGNet \cite{Karen2014}, GoogLeNet \cite{GoogLeNet}, ResNet \cite{ResNet}. 

\subsection{Adversarial attacks}
With the beginning of the integration of ML and DL methods in facial recognition, techniques have also been developed to threaten and to lead to failure of the systems that use these methods. Quite a lot of techniques have been developed which deceive ML models not only by changing their operation but by modifying the inputs to change the factors that the model takes into account when analyzing them. The development of these techniques has been studied extensively both from the analytical point of view and from that of their practical implementation and efficiency in realization in \cite{Ren2020}. Methods have also been developed to detect and prevent these attacks by teaching models through examples of how to spot a real input and a modified one by the attacker and not fall prey to scams \cite{Massoli2019}.

Attacks such as Adversarial Attacks have room for use in many delicate areas which can bring fatal results. With the daily publication of content as textual as well as those in the form of images on the Internet, many machine learning instances have been developed to deal with the evaluation of this content to avoid threatening or uncensored content. Adversarial examples can be used to trick models into faking the contents of photos, which can be used to promote child pornography or spread images with threatening content, without being detected by systems and without being censored. One such system which is exposed to adversarial attacks is content classification on Instagram.

According to \cite{Yuan2017} adversarial attacks are divided into many different types depending on their characteristics. These attacks based on prior knowledge about the system are divided into:
\begin{itemize}
  \item White-box attacks - the attacker has access to the ML model, including the model structure and its parameters.
  \item Black-box attacks - the attacker has no access to the model, but only to its results.
\end{itemize}

Adversarial attacks based on the intent of the attacker can belong to two main types \cite{Yuan2017}:
\begin{itemize}
  \item Targeted attack - develops algorithms that modify the input in such a way that the output is a predefined value, different from the correct one.
  \item Untargeted attack - develops algorithms that modify the input in such a way that the output has a different value from the correct one, but not a predefined value.
\end{itemize}

In \cite{Goodfellow2014}, methods for generating perturbation are presented that modify the input images of the model and deceive it. It explains how models that are easily optimized can be easily fooled as well. Linear models lack the ability to resist perturbation in images, and only structures with a hidden layer need to be trained to resist adversarial patterns.
According to \cite{Massoli2019} adversarial examples are a serious threat to DL models, as they place a significant constraint especially on the use of learning models in sensitive applications where such a threat should be avoided. In addition to adversarial training, one approach that enhances the resilience of AI-based systems to such threats is the detection of intrusive inputs. The adversarial attack on systems that use ML methods to recognize people's faces on the basics learned from similar work will be addressed in the case study.

In \cite{Goswami2018}, experimental evaluation demonstrates that the performance of deep learning based face recognition algorithms can suffer greatly in the presence of by commonly observed distortions in the real world that are well handled by shallow learning methods along with learning based adversaries. That deep CNNs are vulnerable to adversarial examples, which can cause fateful consequences in real-world face recognition applications with security-sensitive purposes is also proved in \cite{Dong2019}.

\cite{DBLPGeorge2019} proposed a multi-channel Convolutional Neural Network based approach for face presentation attack detection (PAD), as well as, authors introduced the new Wide Multi-Channel presentation Attack (WMCA) database for face PAD. The proposed method was compared with feature-based approaches and found to outperform the baselines.

Furthermore, \cite{app10238547} proposed a new framework of palmprint false acceptance attack with a 
deep convolutional generative adversarial network (DCGAN). It has been proven that DCGAN provides better accuracy and significant improvements over the existing classical methods. 
\section{Fast Gradient Sign Method - FGSM} \label{sec:fgsm}

According to \cite{Goodfellow2014}, in many cases the accuracy of an individual input feature is limited. An example of this is the fact that digital images often use only 8 bits per pixel which makes them ignore the information below the dynamic range 1/255. Because the precision of this feature is limited, it is not rational for a classifier(model) to respond differently to an $x$ entry and a modified entry $x^{\prime} = x +\eta$ every element of the $\eta$ is smaller than the precision of the feature. For an example with well-separated classes, we expect the model to show the same class for both $x$ and $x^{\prime}$ while $||\eta||_\infty<\varepsilon$ If we consider the scalar product of a weight vector w and an adversarial example $x^{\prime}$ as follows:

\begin{equation} \label{eq:formulaEpare}
    w^{T}x^{\prime} = w^{T}x + w^{T}\eta
\end{equation}
adversarial perturbation causes activation to increase by $w^{T}\eta$. This increase can be maximized at the maximum rate limit in $\eta$, expressing $\eta$ as $\eta=sign(w)$. If $w$ has $n-$dimensions and the average size of a vector element $w$ is $m$, then the activation would increase by $\varepsilon mn$. According to this flow, a linear model would be forced to follow the signal that most closely matches its weights even if multiple signals are present and the other signals have larger amplitudes.
The purpose of the FGSM method is to add noise to the image whose direction is the same as the gradient of the loss function in relation to the input data analyzed at the pixel level. FGSM has been used as the targeted and untargeted method and the mathematical difference is explained below.

\subsection{Targeted FGSM}
%\hfill\\
The targeted FGSM method, which aims to modify the image so that the model has the desired output, defines the modified image based on the formula:

\begin{equation} \label{eq:formulaEdyte}
\begin{aligned}
   adversarialImage = cleanImage -  \varepsilon sign(\nabla_{cleanImage} \\J(\theta, cleanImage, label))
\end{aligned}
\end{equation}
\newline
where, adversarialImage is the modified image, cleanImage is the clean image to be modified of the tensor format (width, length, depth), the label identifies the class which is intended to be the output value when the model receives the modified image input and is different from the correct output of the model when the input is the pure image, $\varepsilon$ is the coefficient on which the amount of noise added to the image depends and is triggered, $J$ is the cross-entropy loss function and $\theta$ are the model parameters.

\subsection{Untargeted FGSM}
%\hfill\\
The method aims to modify the image so that only if the model has an incorrect output, the modified image is defined by the formula:

\begin{equation} \label{eq:formulaEtrete}
\begin{aligned}
   adversarialImage = cleanImage + \varepsilon sign(\nabla_{cleanImage}\\ 
   J(\theta, cleanImage, label))
\end{aligned}
\end{equation}
\newline
where, the label in this case is the correct output of the model when the input is the clean image.
In both types of methods the noise that is added to the image is defined by the following formula:

\begin{equation} \label{eq:formulaEkatert}
   \eta = \varepsilon sign(\nabla_{cleanImage} J(\theta, cleanImage, label))
\end{equation}
\newline
Creating a finite optimal perturbation linearizes the loss function around the values of the parameters $\theta$. As discussed in \cite{Goodfellow2014}, the fact that such simple and inexpensive algorithms can generate erroneously classified examples serves as evidence in favor of interpreting adversarial examples as results of linearity. Algorithms are useful to accelerate adversarial loss training to analyze trained networks.

% use section* for experimental setup
\section{Experimental Setup}\label{sec:experimental_setup}

\begin{figure*}[t]
     \centering
\includegraphics[width=0.9\textwidth]{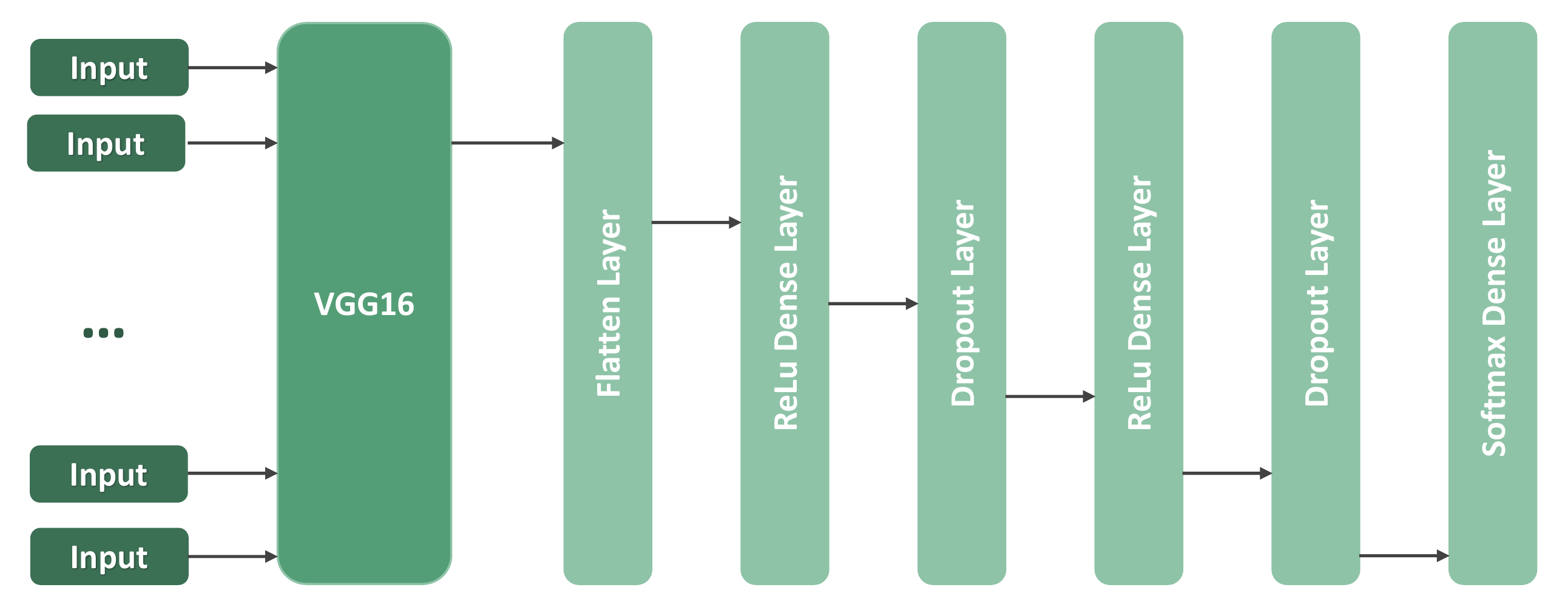}
     \caption{Model architecture}
     \label{fig:model_architecture}
\end{figure*}
Since the application of the attack requires the creation and training of a model with machine learning techniques through artificial neural networks and the most suitable network for training models aimed at classifying images, recognizing images or even detecting objects is the convolutional neural network (CNN), the steps to be followed to achieve training should be devised. 

\subsection{Creating the model for authentication}
%\textcolor{red}{Shto  pershkrim te section}
\subsubsection{Data Collection}

To train the artificial neural network, the model, it is necessary to possess a large set of data, in our case pictures of different persons to have the appropriate relevant material from which the model can learn. The sie of the dataset depends on the classification task and the built up model. CNN models require datasets with thousands of images, but since we use transfer learning the dataset can be much smaller. The dataset we used is the public \emph{"5 Celebrity Faces Dataset"} \footnote{https://www.kaggle.com/dansbecker/5-celebrity-faces-dataset} \cite{dataset} which contains pictures of 5 different people. This is a small database dedicated to experiment with computer vision techniques. For each person in the files, there are 14-20 pictures, structured in respective folders, while in the evaluation files there are 5 photos for each. 

Two main techniques that are used during data collection and preparation are:

\begin{itemize}
  \item Data processing which refers to transformations that must be applied to data before they can be used. Data processing is a technique used to convert raw data into a clean data set. In other words, whenever data is collected from different sources it is collected in raw formats, which is not possible for analysis.
  \item Data augmentation process which is used to add data by transforming current data without applying new data collection. This process is very useful in DL because in cases when it is necessary to collect a very large amount of data, adding them in such a form is very efficient. It enables better learning due to the increase of the training dataset and allows the algorithm to learn from different states. The main operations of this process that we have used are: scaling (ensures that the input is in the range [0, 1]), rotation, zoom, shift, and flip.
\end{itemize}

\vspace{6pt} 
\subsubsection{Defining model architecture}
%\hfill\\
The architecture of the model should be defined by specifying the layers to which the data should be passed to ensure the most efficient training of the model. The specific characteristics of each neural network model are the number and type of layers, the number of neurons, and the type of activation functions. These characteristics must be tested and assigned so that the performance of the model is satisfactory and achieves the level of acceptability for the function it is to perform. Due to the limitation of the topic in analyzing attacks on the authentication process based on facial recognition features, the type of neural network that has been elaborated and implemented is CNN using the convolution operation. The advantage of CNN is obvious considering that it is dedicated to the problems related to computer vision and our related problem.

As documented in \cite{Karen2014}, VGGNet is the CNN network architecture proposed by Karen Simonyan and Andrew Zisserman from Oxford University in 2014. The VGG network inputs are RGB images with 224x224 dimensions. In the VGG16 type architecture, there are 3 fully connected layers and 13 constituent layers, while the VGG19 type has 16 such layers. There are small filters with 3x3 dimensions but with more depth instead of large filters. In both types of architecture, two fully connected layers have 4096 channels while the third layer has 1000 channels to predict 1000 labels. The last fully connected layer uses the softmax activation function to perform the classification. The VGG16 model with pre-trained weights in ImageNet was used as the base model for our trained model. This model supports both image formats with color channel dimensions as \textit{"channels\_first"} and \textit{"channels\_last"}.

% first newpage: goes from pg1:col2 to pg2:col1
Our model, as presented in Figure \ref{fig:model_architecture}, utilizes the features of the VGG16 model and develops on it. The constructed model is a sequential model which by its name is understood to be a sequential set of layers. The number of layers is set so that the network is strong enough to solve the problem for which it is created. Each layer accepts a tensor as input and also has an output tensor. The output of the previous layer is the input of the next layer and by applying this idea to our model we would have a functional structure as follows:
\begin{equation*}
\begin{aligned}
   softmax\_dense\_layer(dropout\_layer (relu\_dense\_layer\\ (dropout(relu\_dense\_layer(flatten\_layer(entry))))))
\end{aligned}
\end{equation*}

\begin{table}[!htb]
  \caption{Output shapes of VGG16 layers}
  \centering
    \begin{tabular}{c c}
    \textbf{Occurrence} & \textbf{Shape}\\ \hline
         & (224, 224, 3) \\
        2 x & (224, 224, 64) \\
         & (112, 112, 64) \\
        2 x & (112, 112, 128) \\
         & (56, 56, 128) \\
        3 x & (56, 56, 256) \\
         & (28, 28, 256) \\
        3 x & (28, 28, 512) \\
        4 x & (14, 14, 512) \\
         & (7, 7, 512) \\ 
    \end{tabular}
    \label{tab:VGG16}
\end{table}

The VGG16 has the input shape as $(224, 224, 3)$ and the output shape as $(7, 7, 512)$. The output shape of each VGG16 layer is represented in Table \ref{tab:VGG16}. Initially the flatten layer is used to flatten the output of the base model VGG16, which was referred to in the previous formula as \emph{entry} and is the input for the flatten layer. The function of the flatten layer is to transform a multidimensional tensor into a one dimensional tensor. Since the output of the previous layer is of shape $(7, 7, 512)$, flatten layer unstacks all the tensor values into a tensor of shape $(7 * 7 * 512)$. In the Dense layer the end result is passed through a fuction called activation function, on our case the used activation function is Rectified Linear Unit (ReLU). This fuction returns 0 for any negative input value and returns the value back for any positive value. This layer specifies the unit value which represents the output size of the layer. The set value is 4096 in both used Dense layers. Droupout layer is used to prevent the model from overfitting. As mentioned in \cite{Srivastava2014} the Neural Networks (NN) with a large number of parameters are very powerful and also slow to use, which makes it difficult to deal with overfitting. Dropout is a technique for addressing this problem by randomly droping units along with their connections, from the NN during training. This prevents units from co-adapting too much. The fraction of the input units to be dropped is set to $0.5$. The last layer is the Dense layer with Softmax activation function, which is used in  classification networks because the result could be interpreted as a probability distribution. This activation function converts the input tensor to a tensor of categorical probabilities, thus dividing the probability of 5 classes. The output shape of each model layer is represented in Table \ref{tab:outputShapes}.

\begin{table}[!htb]
  \caption{Output shapes of the model layers}
  \centering
    \begin{tabular}{c c}
    \textbf{Layer} & \textbf{Shape}\\ \hline
       VGG16 & (7, 7, 512)\\ 
       Flatten & (25088) \\
       Relu Dense & (4096) \\
       Dropout & (4096) \\
       Relu Dense & (4096) \\
       Dropout & (4096) \\
       Softmax & (5) \\
    \end{tabular}
    \label{tab:outputShapes}
\end{table}

\subsubsection{Model training and evaluation}
%\hfill\\
To measure the performance of the classification model we use cross-entropy loss. Cross-entropy loss increases as the predicted probability diverges from the actual label. Since we have more than two labels to predict, it is required to use categorical crossentropy. Optimization algorithms in ML aim at minimizing the loss function. We use RMSprop optimizer on our model. Model training through the epochs has lasted about $100s$ per epoch because image processing requires more time and the structure of the model layers is more complicated. At the end of each epoch, the levels of accuracy and losses during training and validation were recorded.

\subsection{FGSM implementation}
The explanation of FGSM in \cite{Karen2014} has been the basis of realization in the case study. Fast Gradient Sign Method is a method which aims at incorrect classification of images. To achieve such a goal, since it is white box attack, one must know the structure of the model. For this reason, it has been explained in advance how the model that we will use as a target of the attack was constructed. Attack methods have been implemented based on formulas \ref{eq:formulaEdyte} and \ref{eq:formulaEtrete}.

% use section* for results
\section{Results}\label{sec:results}
\begin{figure}
     \centering
     \begin{subfigure}[b]{0.45\textwidth}
         \centering
         \includegraphics[width=0.92\textwidth]{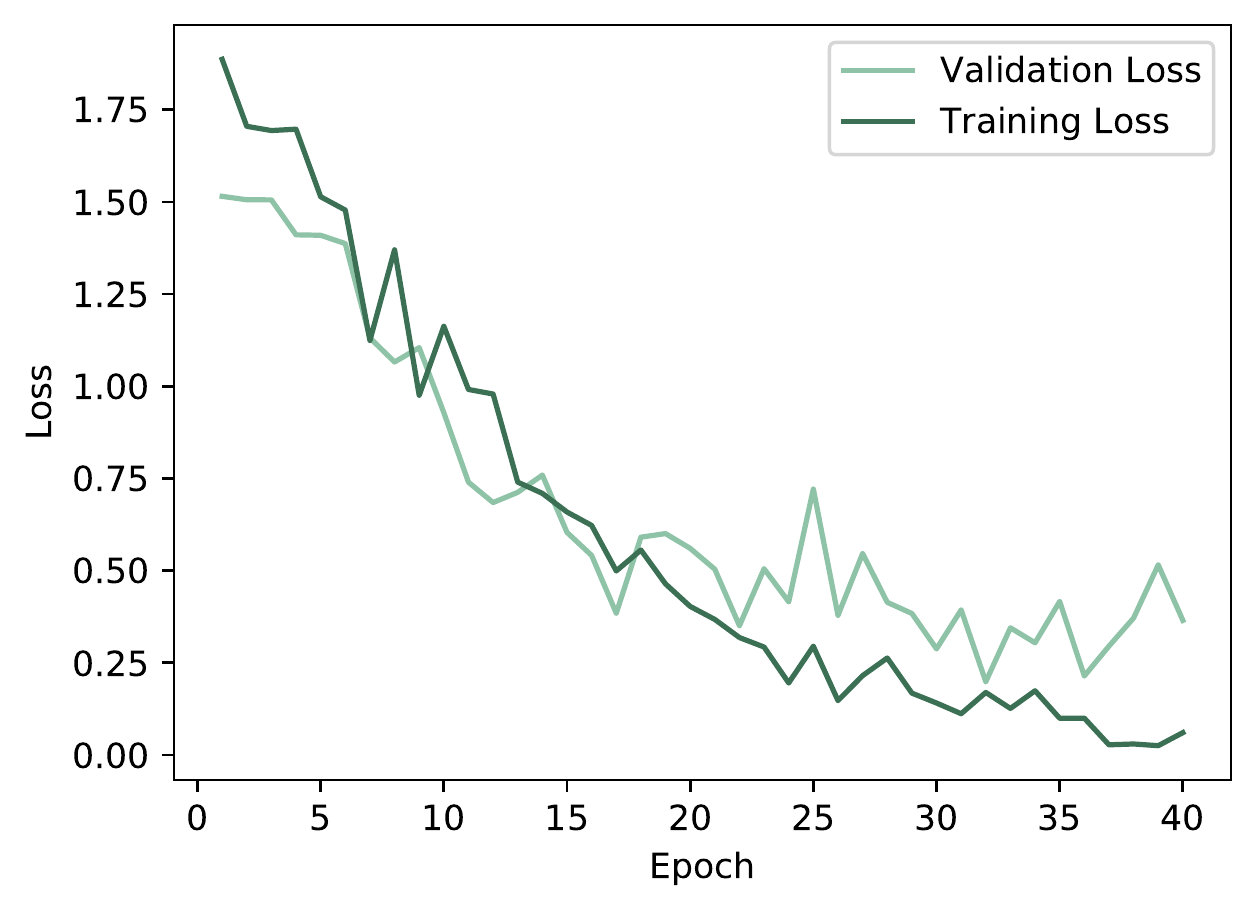}
         \label{fig:lossess}
     \end{subfigure}
     \hfill
     \begin{subfigure}[b]{0.45\textwidth}
         \centering
         \includegraphics[width=0.92\textwidth]{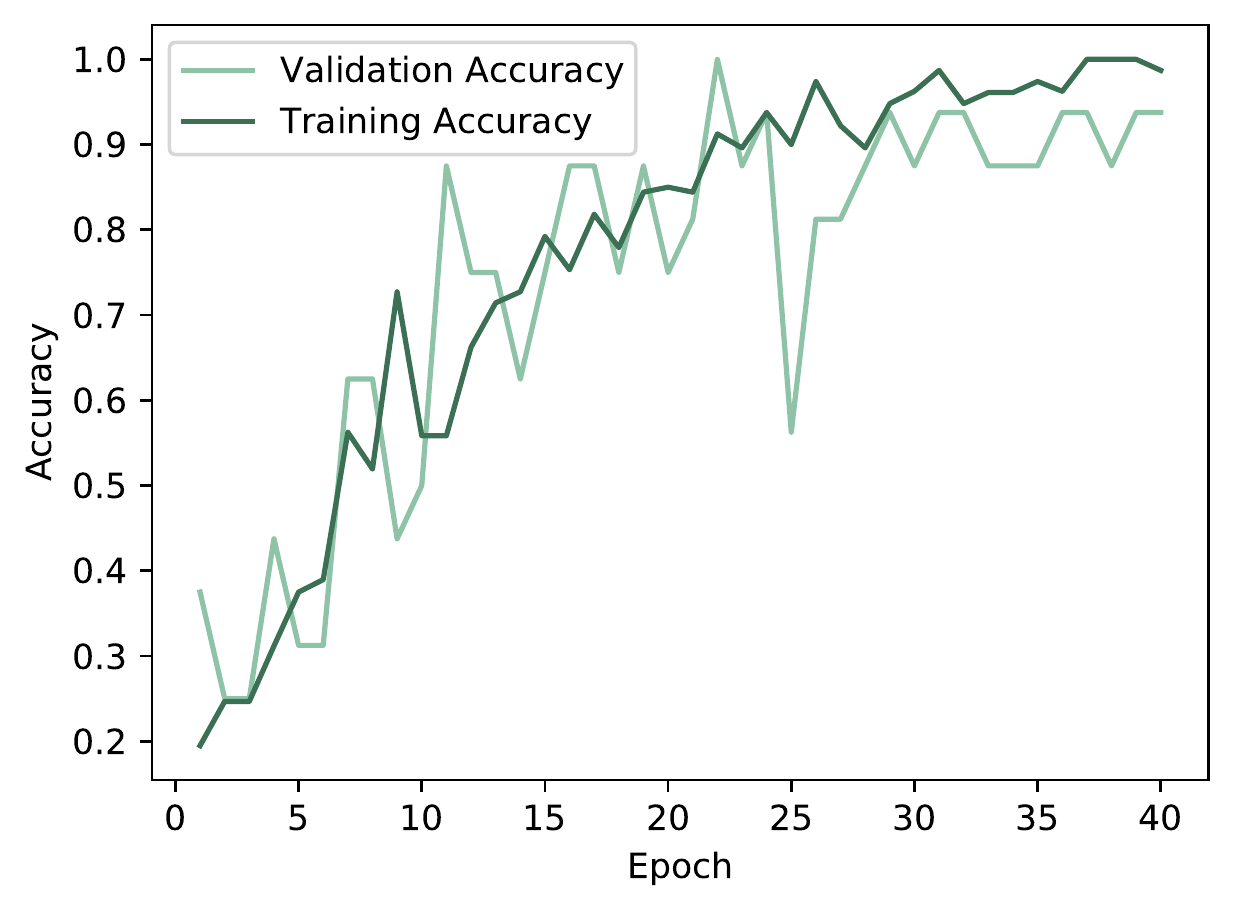}
         \label{fig:graph_res}
     \end{subfigure}
        \caption{The graphical presentation of the losses and accuracy of the proposed model.}
        \label{fig:graphical_results}
\end{figure}

The constructed model we used was trained according to the steps outlined and several tests were performed to observe changes in accuracy and vulnerability spaces when training characteristics change. Five cases have been tried, when the model is trained with $10, 20, 30, 40$ and $50$ epochs. By applying the FGSM method to create modified images, cases have been tried where the value of $\varepsilon$ has changed from the range $0.01, 0.02, 0.03, 0.04$ and $0.05$ and always taking into account whether the method is targeted or untargeted. Thus, five tests were performed with the change of parameters both in terms of training and in terms of attacking the model, where in each test as input were used 5 identically randomly selected images, a photo of each of five people for whose identification the model is trained. A graphical presentation of the losses and accuracy of the model, using 40 epochs, during the training and validation phase is presented in the Figure \ref{fig:graphical_results}. The spike in Figure \ref{fig:graphical_results} shows that the better performance is achieved with 20 and 30 epochs rather than with 25 epochs.
Moreover, these tests have served to build the relationship between the parameters and the impact that their change has on the efficiency of applying the FGSM method in the corresponding model.

%  \begin{figure}[th!]
%      \centering
% \includegraphics[width=0.45\textwidth]{graphics/graph_result.pdf}
%      \caption{Graphical results}
%      \label{fig:graph_results}
%  \end{figure}
 
% \begin{figure}
%      \centering
%      \begin{subfigure}[b]{0.4\textwidth}
%          \centering
%          \includegraphics[width=\linewidth]{graphics/graph_result.pdf}
%          \caption{$y=x$}
%          \label{fig:y equals x}
%      \end{subfigure}
%      \hfill
%      \begin{subfigure}[b]{0.4\textwidth}
%          \centering
%          \includegraphics[width=\linewidth]{graphics/losses.pdf}
%          \caption{$y=3sinx$}
%          \label{fig:three sin x}
%      \end{subfigure}
%         \caption{Three simple graphs}
%         \label{fig:three graphs}
% \end{figure}

From the test results presented in Table \ref{tab:results} one can conclude that the performance of a model which aims to identify and classify five different people depends on the number of epochs assigned for training. The model that has had the highest degree of accuracy is the one trained with $50$ epochs, which has achieved an average of $80.88\%$ confidence in estimation. Hence, this comparison is made by calculating the average level of security accumulated by a large number of test inputs as it is presented as a differentiating value given that other parameters such as maximum security are not efficient at this level as in all three cases when the model trained in $30, 40$ and $50$ epochs, the maximum safety achieved at the exit is $100\%$. 
\begin{table}[htb]
  \caption{Model test results}
  \centering
    \begin{tabular}{c c c c}
    \textbf{Epochs}   & \textbf{Type} & \textbf{Epsilon} & \textbf{Accuracy} \\ \hline
      10 & Targeted & 0.05 & 58.90\% \\ 
      20 & Targeted & 0.04 & 62.33\% \\
      30 & Targeted & 0.01 & 85.17\% \\
      40 & Targeted & 0.04 & 91.78\% \\
      50 & Targeted & 0.03 & 98.38\% \\
      10 & Untargeted & 0.04 & 38.61\% \\
      20 & Untargeted & 0.03 & 92.58\% \\
      30 & Untargeted & 0.03 & 84.02\% \\
      40 & Untargeted & 0.02 & 90.36\% \\
      50 & Untargeted & 0.04 & 89.62\% \\
    \end{tabular}
    \label{tab:results}
\end{table}
Statistically, the cases when the method failed to generate modified images were collected and those images did not represent the desired result from the model. From the analysis of the attack efficiency the highest accuracy is achieved in the application of the untargeted attack in the trained model with $10$ epochs, where it has not resulted in any case of failure and the accuracy of the attack measured with the test samples is $100\%$. Such a percentage of accuracy is evident even in the untargeted attack on the 50-epochs trained model. In both cases the successful or unsuccessful execution of the attack has not been affected by $\varepsilon$, but it can only be argued that its small values has generated noise that has not changed the images much. With the increase of $\varepsilon$ from $0.03$ to $0.05$ in most cases it can be noticed that the picture does not have the original quality. Attacks that have had the highest level of inaccuracy are the targeted attack on the $30$ epochs trained model that achieves $44\%$ inaccuracy and the targeted attack on the $40$ epochs trained model that achieves $36\%$ inaccuracy. In addition to achieving the goal of deceiving a model through modified inputs to have the desired output, the level of security that these unrealistic outputs have should also be discussed. The lowest level of security of the model output evaluation was measured in the unsaturated attack on the trained model with $10$ epochs and with a value of $\varepsilon = 0.01$ with a certainty of $37.32\%$, while the highest security value is recorded in the unsaturated attack on trained model with $40$ epochs and with value of $\varepsilon = 0.02$ that has $90.36\%$ certainty.

From the performed tests it is noticed that there are fluctuations of accuracy in different numbers of epochs. This happens for two reasons, the first because with the change of the number of model epochs the way model fits also changes, facing overfitting or underfitting, while the second is the determination of attack characteristics as epsilon. The ideal number of epochs turned out to be the one that had the highest accuracy, reflecting the change in the accuracy of the adversarial attack.
\section{Conclusion} \label{sec:conclusion}
The FGSM method was mathematically clarified, which was successfully implemented in the case study. From the test results one can conclude that practically this method had a relatively high percentage of accuracy and had the right effect in most cases. 

The results showed that untargeted attacks are more effective and have better performance, with fewer failures and a higher value of security accuracy in estimation. Such an analysis leads us to two choices of adequate conditions. The most effective targeted attack is the trained model's attack with $20$ epochs and $\varepsilon = 0.04$. In comparison, the most effective non-targeted attack is the attack on the trained model with $40$ epochs and $\varepsilon = 0.02$. Achieving such percentages of security in image identification is entirely satisfactory when viewed from the application and well-functioning ML algorithms. Still, the amount of space created for erroneous classifications in delicate processes is intolerable.

Finally, the achieved results could be used in any biometric aware application, with  facial images in the database, where security and authentication is a primary issue, such as access to high security facilities of any kind. 

%------------------------------------------------

\phantomsection
\section*{Acknowledgments} % The \section*{} command stops section numbering

\addcontentsline{toc}{section}{Acknowledgments} % Adds this section to the table of contents

The authors would like to thank the Department of Computer Engineering from the University of Prishtina and the Department of Informatics at the University of Oslo for their support and cooperation.

%----------------------------------------------------------------------------------------
%	REFERENCE LIST
%----------------------------------------------------------------------------------------
\newpage
\phantomsection
\bibliographystyle{unsrt}
\bibliography{Musa_et_al_Author_Version_ArXiv.bib}

%----------------------------------------------------------------------------------------
\end{document}